\documentclass[conference]{IEEEtran}
\IEEEoverridecommandlockouts
\usepackage[numbers,sort&compress]{natbib}
\usepackage{subfigure}
\usepackage{graphicx}
\usepackage{color}

\renewcommand\figurename{Figure}
\renewcommand\thetable{\arabic{table}}
\newcommand{\tabincell}[2]{\begin{tabular}{@{}#1@{}}#2\end{tabular}}

\begin{document}

\title{DeepWriter: A Multi-Stream Deep CNN for Text-independent Writer Identification}

\author{\IEEEauthorblockN{Linjie Xing$^{1,2}$, Yu Qiao$^{1,3}$\IEEEauthorrefmark{1}
\thanks{Yu Qiao is corresponding author.}
}
\IEEEauthorblockA{$^1$Shenzhen key lab of Comp. Vis. \& Pat. Rec., \\
Shenzhen Institutes of Advanced Technology, CAS, China\\
$^2$University of Chinese Academy of Sciences, China   $^3$The Chinese University of Hong Kong, Hong Kong\\
\{lj.xing,yu.qiao\}@siat.ac.cn}}

\maketitle

\begin{abstract}

Text-independent writer identification is challenging due to the huge variation of written contents and the ambiguous written styles of different writers.  This paper proposes \emph{DeepWriter}, a deep multi-stream CNN to learn deep powerful representation for recognizing writers.  \emph{DeepWriter} takes local handwritten patches as input and is trained with softmax classification loss.  The main contributions are: 1) we design and optimize multi-stream structure for writer identification task; 2) we introduce data augmentation learning to enhance the performance of \emph{DeepWriter}; 3) we introduce a patch scanning strategy to handle text image with different lengths.  In addition, we find that different languages such as English and Chinese may share common features for writer identification, and joint training can yield better performance.  Experimental results on IAM and HWDB datasets show that our models achieve high identification accuracy: $99.01\%$ on 301 writers and $97.03\%$ on 657 writers with one English sentence input, $93.85\%$ on 300 writers with one Chinese character input, which outperform previous methods with a large margin.  Moreover, our models obtain accuracy of $98.01\%$ on 301 writers with only 4 English alphabets as input.

\end{abstract}

\section{Introduction}
This paper addresses the problem of automatic writer identification using off-line handwritten images.  Handwriting is a kind of \emph{behavioural biometrics}.  Writer can be recognized by capturing specific characteristics of handwriting habbit of one author, which differ from other authors. \cite{Textural}  Writer identification has been applied in anti-crime and historic document analysis fields, which requires high level of domain expertise and heavy work.

Automatic writer identification aims to recognizing person based on his or her handwritten text.  Researches in writer identification can be divided into two categories, off-line and on-line identification.  On-line writer identification requires record the whole procedure of writing with special devices, thus the input is a time series of pen-tip positions, pressures, angles and other information about writing.  On the other hand, off-line identification merely takes scanned images of handwritten text as input, which is usually more difficult \cite{DeepWriterID}. 

\begin{figure}
\centering
\subfigure[Two English text lines written by writer 009 from IAM dataset]{
\includegraphics[width=8cm]{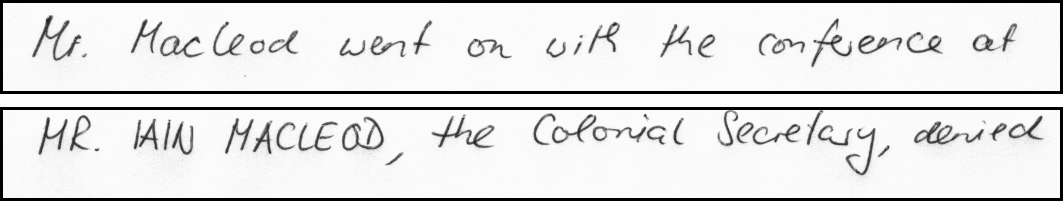}
}
\subfigure[Two English text lines written by writer 010 from IAM dataset]{
\includegraphics[width=8cm]{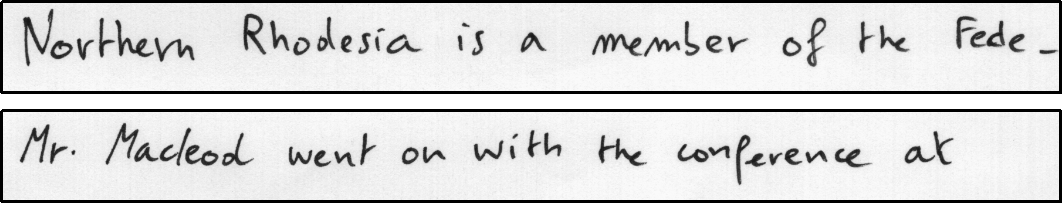}
}
\caption{Different writer examples from IAM dataset}
\label{fig:iam handwritten examples}
\end{figure}

\begin{figure}
\centering
\subfigure[Two Chinese characters written by writer 1001 from HWDB dataset]{
\includegraphics[width=5cm]{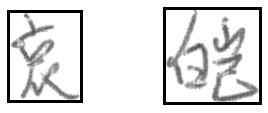}
}
\subfigure[Two Chinese characters written by writer 1002 from HWDB dataset]{
\includegraphics[width=5cm]{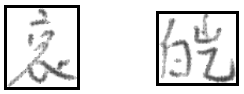}
}
\caption{Different writer examples from HWDB dataset}
\label{fig:hwdb handwritten examples}
\end{figure}

Methods for off-line writer identification can be further categorized into two groups: text-dependent and text-independent.  Text-dependent methods \cite{text-dependent-1,text-dependent-2,text-dependent-3,text-dependent-4} require input image with fixed text contents and which usually compares the input with registered templates for identification.  In contrast with this, text-independent methods \cite{Textural, K-Adjacent, Tan T} dose not make assumptions on input content and have broader applications.  However, compared with text-dependent one, text-independent writer identification needs to deal with image with arbitrary texts which exhibits huge intra-category variations, therefore, and is much more challenging. Figure \ref{fig:iam handwritten examples} and Figure \ref{fig:hwdb handwritten examples} shows several examples of handwritten English and Chinese by different writers. As can be seen, the main difference between two handwritten images is dominated by the text contents.  For writer identification, one needs to extract abstractive written style features and fine details which reflect personal writing habits.  This poses a great challenge for current handcrafted features which usually capture the local shape and gradient information.  These handcrafted features may include both information of written contents (text) and written styles (person), which may limit their performance on this task.

To address this challenging problem, this paper leverages deep CNNs (Convolutional Neural Network) as a powerful model to learn effective representations for off-line text-independent writer identification.  Deep CNNs have demonstrated its effectiveness in various computer vision problems by improving state-of-the-art results with a large margin, including image classification \cite{AlexNet, VGGNet, ResidualNet}, object detection \cite{fast-rcnn, faster-rcnn}, face recognition \cite{deepid, deepid-3}, handwriting recognition  \cite{Character-1} etc.  We propose \emph{DeepWriter}, a multi-stream CNN, for extracting writer-sensitive features. \emph{DeepWriter} takes multiple local regions as input and is trained with softmax loss on identification.  The main contributions are three-folds.  Firstly, we design a multi-stream structure and optimize its configuration for writer identification task.  Secondly, we introduce data augmentation to enhance the performance of DeepWriter.  Finally,  we introduce a patch scanning strategy to handle handwritten image with various lengths.  We evaluate the proposed methods on IAM dataset \cite{IAM} and HWDB1.1 dataset \cite{HWDB}.  Our methods achieves high identification accuracy of $99.01\%$ on 301 writers, $97.3\%$ on 657 writers from the IAM dataset on English sentence level, and $93.85\%$ on 300 writers from HWDB1.1 dataset on Chinese character level, which outperforms previous state-of-the-art. Interestingly, our results also show that handwritten texts of different languages such as English and Chinese may share common features for writer identification, and pretraining CNNs on another language can lead to better performance.

\section{Related Works}
Writer verification is similar to writer identification.  Writer verification system \cite{verification-1, verification-2, verification-3, Textural} performs one-to-one comparison and determines whether or not the two input example are written by the same writer.  Writer identification system \cite{Textural, K-Adjacent} performs a one-to-many search in a large database with handwriting samples of known authorship and returns a likely list of candidates.  Writer verification performs two-class classification, while writer identification performs multi-class classification.  \cite{how much handwritten text} investigates how much handwritten text is needed for text-independent writer verification and identification.  Experimental result in \cite{how much handwritten text} demonstrates that, given the same number of handwritten characters, verification systems achieve lower error rate than identification systems with identical feature.  Therefore, writer identification system is more ambiguous and difficult.

Methods proposed previously generally follow the pipeline of pre-processing, feature extraction and feature matching or classification, and mainly focus on feature extraction.  In \cite{Textural}, Bulace et.al. combined multiple features (directional, grapheme, and tun-length) and used probability distribution functions (PDFs) extracted from the handwriting images to characterize writer individuality, achieving an identification accuracy of $89\%$ on 650 writers from IAM dataset on page level. In \cite{K-Adjacent}, Jain et.al. used K-adjacent segments (KAS) features to model character contours, achieving an identification accuracy of $93.3\%$ on 300 writers from IAM dataset on page level. These methods depend on features defined by humans, which has been shown can be learned automatically by deep CNN.  We believe that with integrated training and overall optimization, deep CNN can learn to extract appropriate features to this task and outperform traditional methods.

\cite{DeepWriterID} leverages CNN to identify writer.  \cite{DeepWriterID} address the problem of on-line text-independent writer identification.  \cite{DeepWriterID} leverages on-line writing information and deep CNNs to obtain accuracy of $95.72\%$ on 187 writers with Chinese page input, and $98.51\%$ on 134 writers with English page input on CASIA Handwriting Database \cite{NLPR}.   In contrast, this paper address the problem of off-line text-independent writer identification which is more general and difficult.  This paper feeds the model with merely scanned gray-scale handwritten image, and learns effective representation with carefully designed deep CNN model, leading a more simplified and elegant method.

\section{DeepWriter}

\begin{figure}
\centering
\includegraphics[width=8cm]{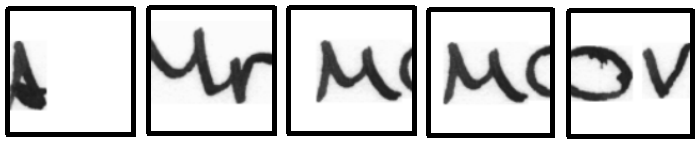}
\caption{Image patches cropped from IAM dataset}
\label{fig:patch examples}
\end{figure}

\begin{figure*}
\centering
\includegraphics[width=18cm]{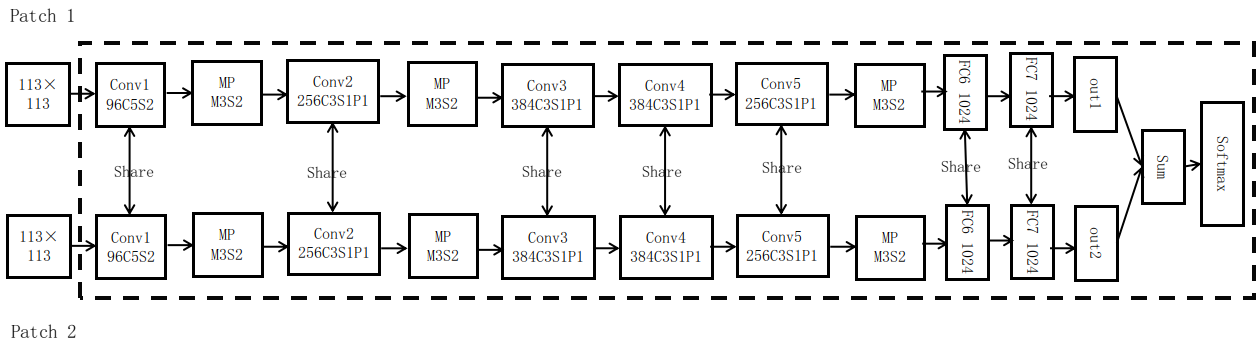}
\caption{Network structure of \emph{DeepWriter}. The boxes with \emph{ConvX} denote convolutional layers.  The $\alpha C \beta S \sigma P \theta$ like notation specifies that the convolutional layer filters the input with $\alpha$ kernels of size $\beta \times \beta$ with a stride of $\sigma$ pixels and a padding of $\theta$ pixels.  The boxes with \emph{MP} denote max-pooling layers.  The $M \beta S \sigma$ like notation specifies that the max-pooling layer performs max-pooling operation in a neighbourhood of size $\beta \times \beta$ with a stride of $\sigma$ pixels.  The boxes with \emph{FCX} denote fully-connected layers, and the followed number specifies the number of neurons.  The \emph{Sum} box denote element-wise sum operation.  The \emph{Softmax} denote softmax classifier.  All convolutional layers and fully-connected layers are followed by Rectified Linear Unit layer(ReLU).  \emph{FC6} and \emph{FC7} are followed by dropout layer with ratio=0.5 to prevents overfitting.}
\label{fig:deepwriter structure}
\end{figure*}

\begin{figure*}
\centering
\includegraphics[width=18cm]{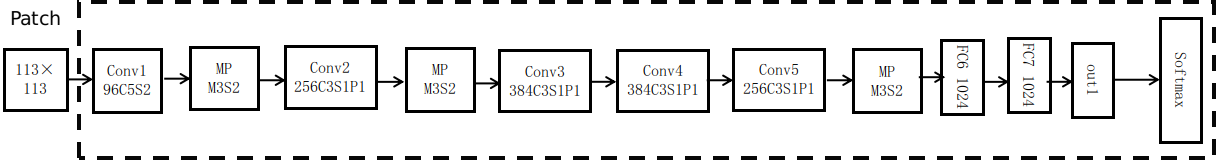}
\caption{Network structure of \emph{Half DeepWriter}}
\label{fig:half deepwriter structure}
\end{figure*}

This section will firstly introduce the design of the multi-stream structure of \emph{DeepWrite} and discuss how to preprocess the input image with various lengths as input for \emph{DeepWrite}.  Then we will describe the training and testing process with implementation details.

\subsection{Multi-Stream}

Our basic network structure is similar to \emph{AlexNet} structure \cite{AlexNet}, as depicted in Figure \ref{fig:half deepwriter structure}.  In this paper, we denote this basic network structure as \emph{Half DeepWriter}.  \emph{Half DeepWriter} takes as input a $113 \times 113$ image patch.  Input handwritten text images for identifying author are with various height and width.  In particular, English sentence handwritten image are usually with high aspect-ratio, whose width is much bigger than its height.  Resizing input image to fixed size distorts the the shape of handwriting, leading serious information loss.  We thus employ a patch scanning strategy to address this problem.  The patch scanning strategy is detailed below.  However, scanning ignores spatial relationships between these image patches, which contains important information to determine the writer.  On the other hand, it is expensive to keep complete spatial relationships between all image patches of input scanned handwritten image.  As a trade-off, we leverage relationship between two adjacent image patches, leading to \emph{DeepWriter} structure.  The network structure of \emph{DeepWriter} is depicted in Figure \ref{fig:deepwriter structure}.  \emph{DeepWriter} takes as input a pair of $113 \times 113$ image patches.  \emph{Patch 2} is adjacent to \emph{Patch 1}, as depicted in Figure \ref{fig:testing pipeline}.  \emph{Out1} and \emph{out2}, output vectors of \emph{FC7} of \emph{DeepWriter}, are merged by element-wise sum operation.  Detailed configuration of \emph{DeepWriter} is specified in the caption of Figure \ref{fig:deepwriter structure}.  The number of model parameters in \emph{DeepWriter} is the same as that in \emph{Half DeepWriter}.  Therefore, \emph{DeepWriter} dose not increase the risk of overfitting, requiring the same size of training data size as \emph{Half DeepWriter}.  We experimentally demonstrate that considering  spatial relationship between image patches benefits writer identification.  The comparison between \emph{DeepWriter} and \emph{Half DeepWriter} on 301 writers from IAM dataset with English sentence handwritten text as input is shown in Table \ref{table:comp. complet and half}.

\begin{table}
\caption{Comparison between \emph{DeepWriter} and \emph{Half DeepWriter}}
\label{table:comp. complet and half}
\centering
\renewcommand{\arraystretch}{1.5}
\begin{tabular}{cc}
\hline
Model & Accuracy \\
\hline
\emph{Half DeepWriter} &  98.23\% \\
\emph{DeepWriter} &  \textbf{99.01\%}\\
\hline
\end{tabular}
\end{table}

\subsection{Patch Scanning Strategy}
Firstly, we resize the image so that min(w,h)=113 while maintaining its aspect ratio.  Secondly, $113 \times 113$ image patches are cropped from the resized image.  Finally, image patches for testing are uniformly sampled from these cropped $113 \times 113$ image patches with a specific ratio.  The sample ratio in this paper is set to 20\% with Chinese character input and 10\% with English sentence input

\subsection{Kernel Size}
\emph{Conv1} and \emph{Conv2} layers of \emph{DeepWriter} and \emph{Half DeepWriter} filter their input with smaller kernels with smaller stride compared to that of \emph{AlexNet}.  This structure adjustment is inspired by the observation that \emph{AlexNet} fed with $131 \times 131$ image patch degrades identification accuracy.  Therefore, we decrease the kernel size and stride step of \emph{Conv1} and \emph{Conv2} layers to handle more image details.  This network structure adjustment also decreases the number of parameters, thus decreasing the risk of overfitting.  The comparison between  \emph{AlexNet} and its variants on 301 writers from IAM dataset with English handwritten image patch as input is shown in Table \ref{table:comp. kernel size}.

\begin{table}
\caption{Kernel size comparison}
\label{table:comp. kernel size}
\centering
\renewcommand{\arraystretch}{1.5}
\begin{tabular}{ccc}
\hline
Patch size & Configuration & Accuracy \\
\hline
$227 \times 227$ & \tabincell{c}{\emph{Conv1}:$96C11S4$ \\ \emph{Conv2}:$256C5S1P2$} & $91.20\%$ \\
$131 \times 131$ & \tabincell{c}{\emph{Conv1}:$96C11S4$ \\ \emph{Conv2}:$256C5S1P2$} & $87.13\%$ \\
$113 \times 113$ & \tabincell{c}{\emph{Conv1}:$96C5S2$ \\ \emph{Conv2}:$256C3S1P1$} & \textbf{91.35\%} \\
\hline
\end{tabular}
\end{table}

\subsection{Neuron Number}
Comparing to \emph{AlexNet}, \emph{FC6} and \emph{FC7} layers of \emph{DeepWriter} and \emph{Half DeepWriter} have less neurons.  The size of training data and number of classes of this task  are smaller than those of ILSVRC \cite{ILSVRC}. Therefore We believe that appropriate neuron number reduces the risk of overfitting.  We chose the number of neurons of \emph{FC6} and \emph{FC7} through contrast experiment on validation set, varying neuron number of \emph{Half DeepWriter}, on 301 writers from IAM dataset with English handwritten image patch as input.  Experiment result is shown in Table \ref{table:comp. neuron number}.  We finally set the neuron number of \emph{FC6} and \emph{FC7} layers of \emph{DeepWriter} and \emph{Half DeepWriter} to 1024.

\begin{table}
\caption{Neuron number comparison}
\label{table:comp. neuron number}
\centering
\renewcommand{\arraystretch}{1.5}
\begin{tabular}{cc}
\hline
Neuron number & Accuracy \\
\hline
4096 & $91.35\%$ \\
1024 & \textbf{92.15\%} \\
512  & $91.10\%$ \\
\hline
\end{tabular}
\end{table}

\subsection{Feature Sharing}
We also observe that handwritten images of different languages share some common features for identifying writers.  On IAM dataset, we finetune \emph{DeepWriter} from \emph{Half DeepWriter} model pretained on HWDB1.1, whose data size is much bigger than IAM dataset.  On HWDB1.1 dataset, we finetune \emph{Half DeepWriter} from the above \emph{DeepWriter} model.  Table \ref{table: comp. feature sharing} shows comparison between whether joint training or not.

\begin{table}
\caption{Benefit from joint training}
\label{table: comp. feature sharing}
\centering
\renewcommand{\arraystretch}{1.5}
\begin{tabular}{ccc}
\hline
Dataset & Train & Accuracy \\
\hline
IAM & Pretrained on HWDB &  \textbf{99.01\%} \\
IAM & Trained directly on IAM &  98.80\% \\
HWDB1.1 & Pretrained on IAM & \textbf{93.85\%} \\
HWDB1.1 & Trained directly on HWDB1.1 & 93.45\% \\
\hline
\end{tabular}
\end{table}

\subsection{Training Details}
We augment training data by resizing the shorter edge of input image to 113 with original aspect ratio and then randomly cropping $113 \times 113$ image patches from the input image.  It is important to keep the original aspect ratio which contains important information of handwriting habits for identifying writer.  The identification accuracy degrades seriously when the input image is distorted.

Firstly, the \emph{Half DeepWriter} was trained on HWDB1.1 dataset.  We trained \emph{Half DeepWriter} using mini-batch gradient descent.  The batch size was set to 256, momentum to 0.9, and weight decay to $5 \times 10^{-4}$.  The learning rate was initialized at $10^{-2}$, and then decreased by a factor of 10 every $10^{5}$ iterations.  The learning was stopped after 400K iterations.

Secondly, the \emph{DeepWriter} for IAM dataset was fintuned from \emph{Half DeepWriter} model pretained on HWDB1.1 dataset.  The batch size was set to 256, momentum to 0.9, and weight decay to $5 \times 10^{-4}$.  The base learning rate was initialized at $10^{-3}$, and then decreased by a factor of 10 every 20K iterations.  The learning was stopped after 40K iterations.  The learning rate of softmax layer correlated to specific dataset was set to tenfold larger than base learning rate.

Finally, the \emph{Half DeepWriter} was finetuned from the above \emph{DeepWriter} model in the same way as that of training directly.

\subsection{Testing Details}
\begin{figure*}
\centering
\includegraphics[width=13.5cm]{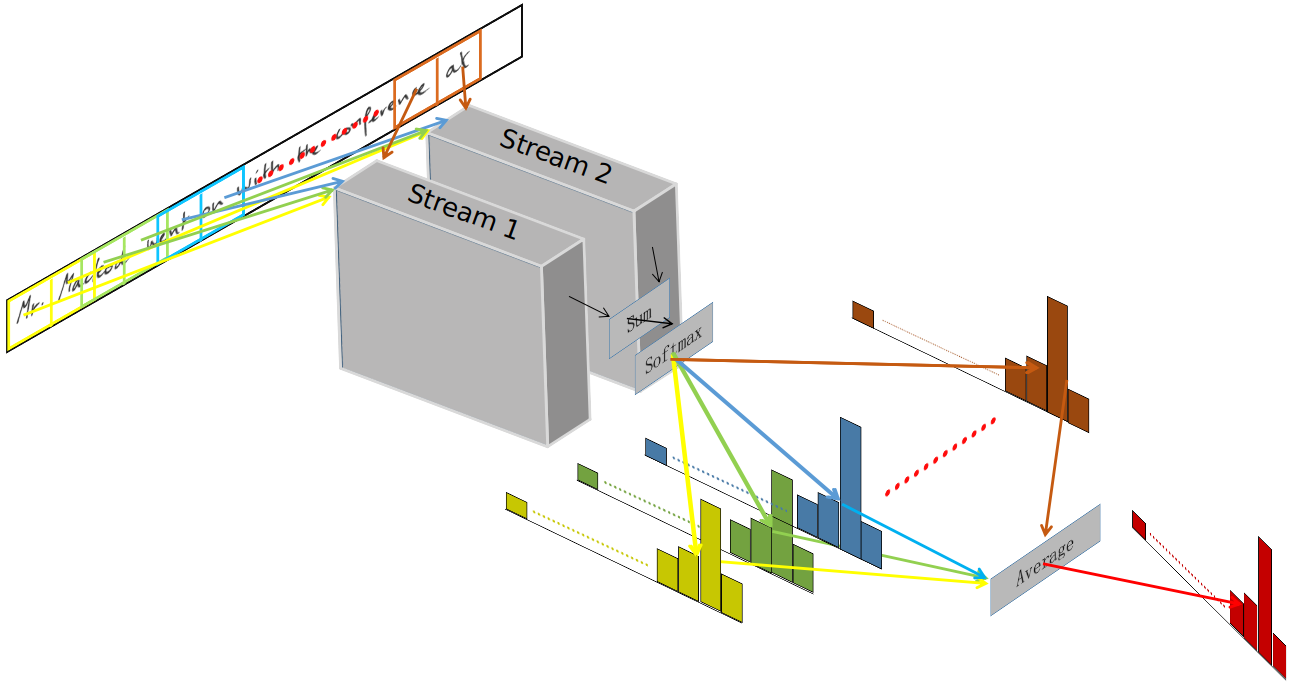}
\caption{Pipeline of testing. Stream 1 and Stream 2 share the same parameters.}
\label{fig:testing pipeline}
\end{figure*}

Given a scanned handwritten image, the testing procedure follows this pipeline: scan the image to generate image patches following the strategy presented above;  input $ith$ image patch pair or image patch into \emph{DeepWriter} or \emph{Half DeepWriter} to compute score vector $f_i$;  compute final score of $jth$ writer $f_{j} = \frac{1}{N}\sum_{i=1}^{N}{f_{ij}}$, where $N$ denotes the number of image patches;  return the writer with highest score.  Noting that the score vector outputted by \emph{DeepWriter} can be treated as a probability distribution over all writers, we thus average score vectors of all image patch pairs or image patch to construct the final prediction of input image.  The testing pipeline is depicted in Figure \ref{fig:testing pipeline}.

\section{Experiments}

\subsection{Data sets}
The IAM dataset (version 3.0) \cite{IAM} contains unconstrained handwritten English text from 657 different writers, using different pens. Handwritten pages in IAM dataset were scanned at a resolution of 300dpi and saved as PNG images with 256 gray levels.  IAM dataset contains 1,539 pages of scanned text which contains 5,685 isolated sentences.  301 writers contribute more than 1 page of scanned text.  In this paper, we train, validate and test in sentence images.  Sentence images contributed by each writer are divided into training set, validation set and testing set according to the ratio 4 : 1 : 1.

The HWDB1.1 dataset \cite{HWDB} contains handwritten Chinese text from 300 different writers, which were scanned at a resolution of 300dpt and saved with 256 gray levels.  HWDB1.1 contains 1,172,907 Chinese character images.  Each writer contributes about 3,755 different Chinese characters.  The Chinese character images contributed by each writer are divided into training set, validation set, and testing set according to the ratio 4 : 1 : 1.

\subsection{Experimental Results}
We use the off-the-shelf resource Caffe \cite{caffe} to train our \emph{Half DeepWriter} and \emph{DeepWriter}.  Our \emph{Half DeepWriter} achieves identification accuracy of $93.85\%$ on 300 writers with merely one Chinese character input.  Our \emph{DeepWriter} achieves identification accuracy of $99.01\%$ on 301 writers from IAM dataset on English sentence level, $97.3\%$ on 657 writers from IAM dataset on English sentence level.  In addition, \emph{DeepWriter} achieves identification accuracy of $96.92\%$ When given two adjacent English handwritten image patches, which usually cover 2 to 3 English alphabets.  \emph{DeepWriter} taking as input three adjacent image patches, which usually cover 3 to 4 English alphabets, achieves identification accuracy of $98.01\%$.  Experimental results above demonstrate that our models can obtain high identification accuracy with little handwritten text input.

We summarize experiment results of our method and several published writer identification methods in Table \ref{table:comparison}.  \cite{Textural, K-Adjacent, Ink-trace-12, Texture-13, Junction-15, Texture-16} follow the classic pipeline to address off-line writer identification problem: propose and combine multiple handcrafted features; employ Euclidean, cosine or trained SVM(Support Vector Machines) as similarity metric; perform nearest neighbour search to compute writer of input handwritten image.  \cite{DeepWriterID} employs Deep CNNs to address on-line writer identification problem, as summarized in RELATED WORKS section.  Our method outperforms previous start-of-art methods a large margin.  \emph{DeepWriter} achieve similar identification accuracy with much less input text.  In addition, \emph{DeepWriter} only need to store the trained model for test, without storing big reference data set.  Because \emph{DeepWriter} dose not need to perform heavy search computation, the test procedure is fast.

\begin{table*}
\renewcommand{\arraystretch}{1.5}
\caption{Comparison with text-independent writer identification methods}
\label{table:comparison}
\centering
\begin{tabular}{cccccccc}
\hline
  &  Year  &  Input type  &  Dataset  &  Language  &  Number of writer  &  Input text for test  &  Accuracy  \\
\hline
\emph{DeepWriter}  &  2016  &  off-line  &  IAM  &  English  &  301  &  1 sentence  &  99.01\% \\
\emph{DeepWriter}  &  2016  &  off-line  &  IAM  &  English  &  657  &  1 sentence  &  97.3\%  \\
\emph{DeepWriter}  &  2016  &  off-line  &  IAM  &  English  &  301  &  about 3 alphabets  &  96.92\%  \\
\emph{DeepWriter}  &  2016  &  off-line  &  IAM  &  English  &  301  &  about 4 alphabets  &  98.01\%  \\
\emph{Half DeepWriter}  &  2016  &  off-line  &  HWDB1.1  &  Chinese  &  300  &  1 character  &  93.85\% \\

\hline

Bulacu \emph{et al.} \cite{Textural}  &  2007  &  off-line  &  IAM  & English  &  650  &  1 page &  89\%  \\
Jain \emph{et al.} \cite{K-Adjacent}  &  2011  &  off-line  &  IAM  &  English  &  300  &  1 page  &  93.3\%  \\
Jain \emph{et al.} \cite{K-Adjacent}  &  2011  &  off-line  &  IAM  &  English  &  650  &  1 page  &  92.1\%  \\
Brink \emph{et al.} \cite{Ink-trace-12}  &  2012  &  off-line  &  IAM  &  English  &  657  &  1 page  &  97\%  \\
Bertolini \emph{et al.} \cite{Texture-13}  &  2013  &  off-line  &  IAM  & English  & 650 & 1 page & 96.7\%  \\
He \emph{et al.} \cite{Junction-15}  &  2015  &  off-line  &  IAM  & English  &  650  &  1 page  &  91.1\%  \\
Hannad \emph{et al.} \cite{Texture-16}  &  2016  &  off-line  &  IAM  &  English  &  657  & 6 text lines at most &  89.54\%  \\
Yang \emph{et al.} \cite{DeepWriterID}  &  2015  &  on-line  &  CASIA Handwriting Database  &  English  &  134  &  1 page  &  98.51\%  \\
Yang \emph{et al.} \cite{DeepWriterID}  &  2015  &  on-line  &  CASIA Handwriting Database  &  Chinese  &  187  &  1 page  &  95.72\%  \\

\hline
\end{tabular}
\end{table*}

\section{Conclusion and Future Work}
In this paper, we introduce a novel data-driven text-independent model to identify writer for off-line handwritten scanned images.  We learn a carefully designed deep Convolutional Neural Network to extract discriminative features from handwritten image patches.  We investigate how the network structure affects identification accuracy and introduce multi-stream structure to leverage spatial relationship between handwritten image patches.  We also investigate the appropriate method to augment training data for writer identification.  We achieve high identification accuracy even merely taking as input one Chinese character or 4 English alphabets.  In the future, we will investigate the off-line text-independent writer verification task with discriminative features extracted by \emph{DeepWriter}.  We will also investigate multi-task learning of identification and verification.

\end{document}